%% file: main.tex
\PassOptionsToPackage{x11names,table,dvipsnames}{xcolor}

\documentclass[acmtog]{acmart}

\AtBeginDocument{%
  \providecommand\BibTeX{{%
    \normalfont B\kern-0.5em{\scshape i\kern-0.25em b}\kern-0.8em\TeX}}}

\copyrightyear{2022}
\acmYear{2022}
\setcopyright{acmcopyright}\acmConference[SIGGRAPH '22 Conference Proceedings]{Special Interest Group on Computer Graphics and Interactive Techniques Conference Proceedings}{August 7--11, 2022}{Vancouver, BC, Canada}
\acmBooktitle{Special Interest Group on Computer Graphics and Interactive Techniques Conference Proceedings (SIGGRAPH '22 Conference Proceedings), August 7--11, 2022, Vancouver, BC, Canada}
\acmPrice{15.00}
\acmDOI{10.1145/3528233.3530727}
\acmISBN{978-1-4503-9337-9/22/08}

\acmSubmissionID{483}

\DeclareMathOperator*{\argmax}{arg\,max}
\DeclareMathOperator*{\argmin}{arg\,min}

\usepackage{tabularx}
\usepackage{booktabs}
\usepackage{tikz}
\usepackage{wrapfig}
\usepackage[percent]{overpic}
\usepackage[binary-units]{siunitx}

\begin{document}

\title{Variable Bitrate Neural Fields}


\author{Towaki Takikawa}
\affiliation{%
  \institution{NVIDIA, University of Toronto}
  \country{Canada}
}
\email{tovacinni@gmail.com}

\author{Alex Evans}
\affiliation{%
  \institution{NVIDIA}
  \country{United Kingdom}
}
\email{alexe@nvidia.com}

\author{Jonathan Tremblay}
\affiliation{%
  \institution{NVIDIA}
  \country{United States of America}
}
\email{jtremblay@nvidia.com}

\author{Thomas Müller}
\affiliation{%
  \institution{NVIDIA}
  \country{Switzerland}
}
\email{tmueller@nvidia.com}

\author{Morgan McGuire}
\affiliation{%
  \institution{ROBLOX, University of Waterloo}
  \country{Canada}
  \email{morgan@roblox.com}
}

\author{Alec Jacobson}
\affiliation{%
  \institution{Adobe Research, University of Toronto}
  \country{Canada}
}
\email{jacobson@cs.toronto.edu}

\author{Sanja Fidler}
\affiliation{%
  \institution{NVIDIA, University of Toronto}
  \country{Canada}
}
\email{fidler@cs.toronto.edu}

\renewcommand{\shortauthors}{Takikawa et al.}

\begin{abstract}
Neural approximations of scalar- and vector fields, such as signed distance functions and radiance fields, have emerged as accurate, high-quality representations.
State-of-the-art results are obtained by conditioning a neural approximation with a lookup from \emph{trainable feature grids}~\cite{liu2020neural, takikawa2021neural, martel2021acorn, muller2022instant} that take on part of the learning task and allow for smaller, more efficient neural networks.
Unfortunately, these feature grids usually come at the cost of significantly increased memory consumption compared to stand-alone neural network models.
We present a dictionary method for compressing such feature grids, reducing their memory consumption by up to $100\times$ and permitting a multiresolution representation which can be useful for out-of-core streaming.
We formulate the dictionary optimization as a vector-quantized auto-decoder problem which lets us learn end-to-end discrete neural representations in a space where no direct supervision is available and with dynamic topology and structure. Our source code is available at \url{https://github.com/nv-tlabs/vqad}.
\end{abstract}

\begin{teaserfigure}
\begin{tikzpicture}
\node (image) at (0,0) {
  \includegraphics[width=\textwidth]{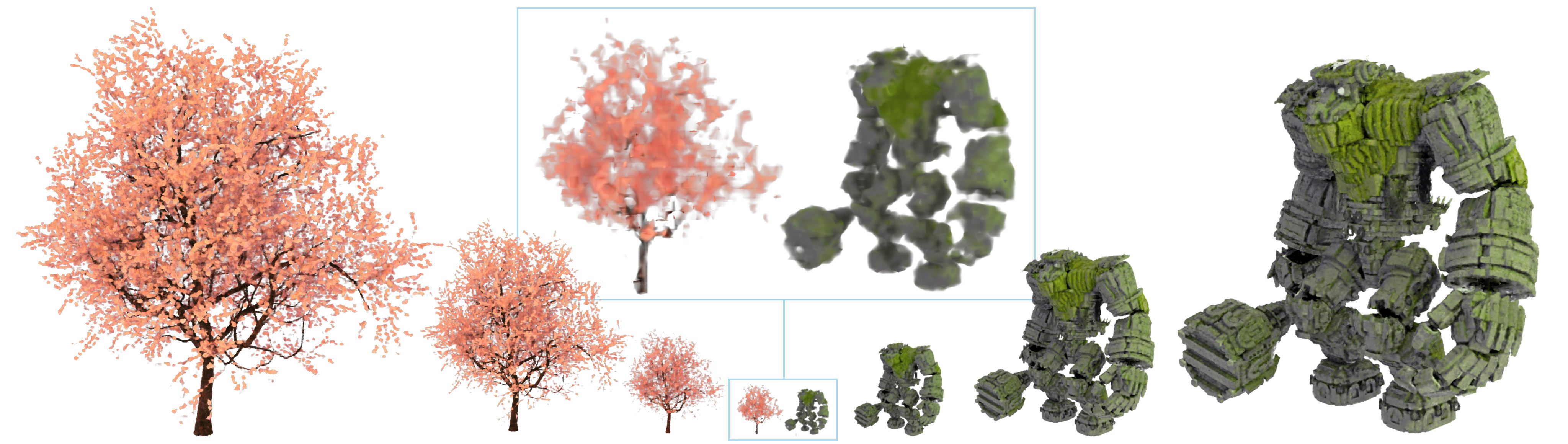}
};
\node[align=center] at ( 6.90,-2.8) {\SI{531}{\kilo\byte}}; 
\node[align=center] at ( 3.30,-2.8) {\SI{131}{\kilo\byte}};
\node[align=center] at ( 1.35,-2.8) {\SI{37}{\kilo\byte}};
\node[align=center] at ( 0.38,-2.8) {\SI{17}{\kilo\byte}}; 
\node[align=center] at (-0.38,-2.8) {\SI{18}{\kilo\byte}}; 
\node[align=center] at (-1.35,-2.8) {\SI{46}{\kilo\byte}};
\node[align=center] at (-3.10,-2.8) {\SI{160}{\kilo\byte}};
\node[align=center] at (-6.65,-2.8) {\SI{563}{\kilo\byte}};
\end{tikzpicture}
  \caption{\textbf{Compressed streaming level of detail.} Using our {\em vector-quantized auto-decoder} (VQ-AD) method, 
  we compactly encode a 3D signal in a hierarchical representation which can be used for progressive streaming and level of detail (LOD). 
  Two example neural radiance fields are shown after streaming from 5 to 8 levels of their underlying octrees. The sizes shown are the total bytes streamed; that is, the finer LODs include the cost of the coarser ones. Prior work such as NeRF \cite{mildenhall2020nerf} requires $\approx$ \SI{2.5}{\mega\byte} to be transferred before anything can be drawn. 
  }
  \label{fig:teaser}
\end{teaserfigure}
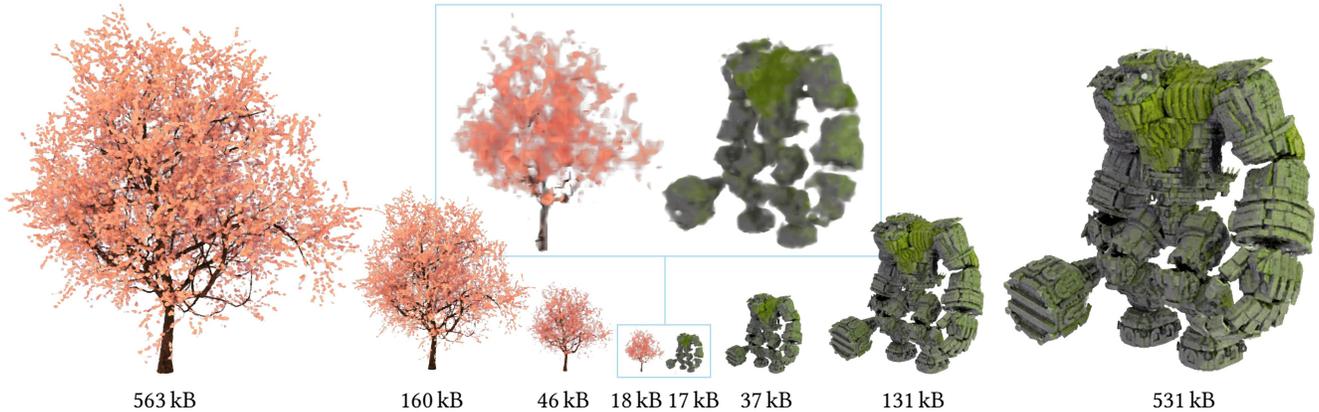

\maketitle

\input{src/1_introduction}
\input{src/2_background}
\input{src/3_method}
\input{src/4_results}
\input{src/5_conclusion}

\begin{acks}

We would like to thank Joey Litalien, David Luebke, Or Perel, Clement Fuji-Tsang, and Charles Loop for a whole lot of fruitful discussion for this project. We would also like to thank Alexander Zook, Koki Nagano, Jonathan Grasnkog, and Stan Birchfield for their help with reviewing the draft for this paper. 

\end{acks}

\bibliographystyle{ACM-Reference-Format}
\bibliography{base}

\input{src/6_supplemental}

\end{document}

%% file: src/1_introduction.tex
\begin{figure}
\begin{tikzpicture}
\node (image) at (0,0) {
        \includegraphics[width=\linewidth]{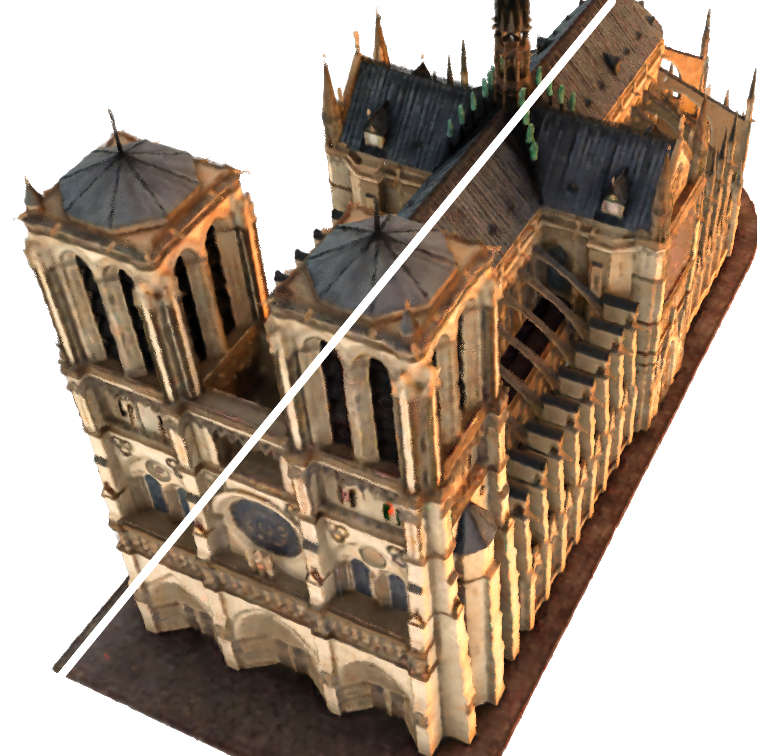}
    };
\node[align=left] at (-3.1,3.5) {\SI{15207}{\kilo\byte} \\ 31.31 PSNR} ;
\node[align=right] at (3.1,-3.0) {\SI{252}{\kilo\byte} \\ 30.20 PSNR} ;

\end{tikzpicture}
  \caption{\textbf{Feature Grid Compression}. 
 Top-left shows a baseline neural radiance field whose \emph{uncompressed} feature grid weighs \SI{15207}{\kilo\byte}. Our method, shown bottom right, compresses this by a factor of 60x, with minimal visual impact (PSNR shown relative to training images). In a streaming setting, a coarse LOD can be displayed after receiving only the first {\SI{10}{\kilo\byte}} of data. All sizes are without any additional entropy encoding of the bit-stream.
 }
  \label{fig:featuregridcompression}
\end{figure}

\section{Introduction}
\label{sec:intro}

Coordinate-based multi-layer perceptrons (MLPs) have emerged as a promising tool for computer graphics for tasks such as view synthesis~\cite{mildenhall2020nerf}, radiance caching~\cite{ren2013global, muller2021real}, geometry representations~\cite{park2019deepsdf,davies2020effectiveness}, 
and more~\cite{neuralfields2021}. 
Whereas discrete signal representations like pixel images or voxels approximate continuous signals with regularly spaced samples of the signal, these \textit{neural fields} approximate the continuous signal directly with a continuous, parametric function,
{\em i.e.}, a 
MLP which takes in coordinates as input and outputs a vector (such as color or occupancy).\@ 

\textit{Feature grid methods}~\cite{liu2020neural, takikawa2021neural, martel2021acorn, Chan2021, muller2022instant} are a special class of neural fields which have enabled state-of-the-art signal reconstruction quality whilst being able to render~\cite{takikawa2021neural} and train at interactive rates~\cite{muller2021real}. These methods embed coordinates into a high dimensional space with a lookup from a parametric embedding (the feature grid), in contrast to non-feature grid methods which embed coordinates with a fixed function such as positional Fourier embeddings~\cite{tancik2020fourier}. This allows them to move the complexity of the signal representation away from the MLP and into the feature grid (a spatial data structure such as a sparse grid~\cite{liu2020neural, takikawa2021neural} or a hash table~\cite{muller2022instant}).
These methods require high-resolution feature grids to achieve good quality. This makes them less practical for graphics systems which must operate within tight memory, storage, and bandwidth budgets. Beyond compactness, it is also desirable for a shape representation to dynamically adapt to the spatially varying complexity of the data, the available bandwidth, and desired level of detail.

In this paper, we propose the \textit{vector-quantized auto-decoder}~(VQ-AD) method to directly learn compressed feature-grids for signals without direct supervision. Our representation enables progressive, \textit{variable bitrate} streaming of data by being able to scale the quality according to the available bandwidth or desired level of detail, see Figure~\ref{fig:teaser}. 
Our method enables end-to-end \textit{compression-aware} optimization which results in significantly better results than typical vector quantization methods for discrete signal compression. 
We evaluate our method by compressing feature-grids which represent neural radiance fields (NeRF)~\cite{mildenhall2020nerf} and show that our method is able to reduce the storage required by two orders of magnitude with relatively little visual quality loss without entropy encoding (see Figure \ref{fig:featuregridcompression}).

%% file: src/2_background.tex
\section{Related Works}

\subsection{Compression for Computer Graphics}

The ability to dynamically compress and filter data is of great importance in computer graphics. 
Nanite~\cite{karis2021nanite} uses mesh levels of detail~\cite{luebke2003level} for out-of-core streaming of assets, which adapts to the image-space footprint to decouple the complexity of assets from the complexity of the render. \citet{balsa2014state} survey techniques to render massive volume datasets, which make extensive use of streaming and compression. \citet{maglo20153d} survey mesh compression, which include progressive streaming methods. Prefiltered representations like mipmaps~\cite{williams1983pyramidal} are heavily utilized in real-time graphics systems.

Most relevant to our paper are the works on the compression and streaming of voxels, which approximate volumes with regularly spaced samples. 
The primary challenge for voxel-based systems is their high memory utilization. \citet{crassin2009gigavoxels} adopt a block-based $N^3$-tree~\cite{lefebvre2005octree}, which stores dense $N^3$ bricks in a sparse tree. They take advantage of the efficient GPU texture unit and leverage lazy out-of-core streaming with cone-tracing~\cite{crassin2011interactive}. These approaches require extra care to handle interpolation at block boundaries, which can either be handled through neighbour pointers~\cite{ljung2006multiresolution} or by allocating extra storage for redundant border elements~\cite{crassin2009gigavoxels}. Filtering data in voxel structures has also been well studied~\cite{heitz2012representing, heitz2015sggx}.
The nodes of such $N^3$-trees can be compressed with transform coding. \citet{tang2018real} compress with a block-wise Karhunen-Loève transform (KLT). \citet{wang2019learned} and \citet{tang2020deep} compress with auto-encoders. \citet{zhang2014point} use a tree whose nodes are sparse blocks encoded with a graph Laplacian transform. \citet{de2016compression} compress with a global transform on the sparse tree structure using an adaptive Haar wavelet transform. Efficient software libraries like OpenVDB~\cite{museth2019openvdb, museth2021nanovdb} exist to work with sparse brick structures. Our work is similar in that we want to compress a feature grid, which consists of sparse blocks of features that condition a neural network; however this comes with additional difficulties of having to optimize with respect to loss functions on downstream tasks. We take inspiration from these works and propose an end-to-end trainable compression scheme for feature grids. By storing data at \emph{all} levels of a multi-resolution octree as in \cite{takikawa2021neural, crassin2009gigavoxels}, we can stream the resulting data structure in a breadth-first fashion. This allows coarse levels of detail to be rendered almost immediately, with features from the subsequent tree levels progressively refining the model.

\subsection{Compression for Neural Fields}

Compression is one of the often mentioned benefits of using neural fields to represent signals, however there are still relatively few works which evaluate this property. We review works which evaluate compression for both global methods which use standalone neural networks, as well as feature-grid methods.

\subsubsection{Global methods} 
Many works~\cite{davies2020effectiveness, dupont2021coin, zhang2021implicit} formulate compression as an architecture search problem where a hyperparameter sweep is used to find the optimal architecture with the desired rate-distortion tradeoff, and variable bitrate is achieved by storing multiple models. \citet{bird20213d} directly minimize the rate-distortion tradeoff with a differentiable approximation of entropy, and variable bitrate is achieved by tuning for different tradeoffs. \citet{lu2021compressive} use vector quantization of MLP parameters alongside quantization at different bitwidths for variable rate compression. These are all global-methods and hence reformulate the problem as neural network model compression. 
Although not for compression, \citet{lindell2021bacon} learn a series of bandlimited signals with multiplicative filter networks~\cite{fathony2020multiplicative}, \citet{barron2021mip} propose a special positional encoding function which can represent the expanding spatial footprint of cone-tracing samples, and \citet{baatz2021nerf} uses the same spatial footprint as an input to the neural network to filter.
The resulting filterable representations have fixed size (static bitrate), in contrast to our work where we aim to achieve variable bitrate via streaming level of detail.

\subsubsection{Feature-grid methods}
\citet{takikawa2021neural} learn a multiresolution tree of feature vectors which can be truncated at any depth; this achieves an adaptive bitrate through streaming of a breadth-first prefix of the tree. \citet{isik2021lvac} directly learn the transform coefficients of an adaptive Haar wavelet transform for a feature grid. \citet{muller2022instant} use a hash table to learn compact but fixed-size feature grids; large tables are required for good quality. We see these methods as complementary to our contributions. 
In all of these works, neural signal compression is generally treated as a separate problem from discrete signal compression. In this paper, we show-case how the auto-decoder framework~\cite{park2019deepsdf} can directly bridge the gap between discrete signal compression and neural signal compression. 

\section{Background}
We will first give a background on signal compression and neural fields to provide an overview of the terminology and concepts used throughout the paper. 
We define a signal $u(x) : \mathbb{R} \to \mathbb{R}$ as a continuous function which maps a coordinate $x$ to a value. In discrete signal processing, signals are typically approximated by a sequence of values of length $n$, representing evenly spaced samples of the continuous signal: 
\begin{equation}
    u_x = [u_1, ...,u_n ]
\end{equation}
where $u_i$ denotes the $i$-th sample in the sequence. 
If the continuous signal $u(x)$ is bandlimited and the spacing of the samples $u_x$ exceeds the Nyquist rate~\cite{shannon1984communication}, the continuous signal can be reconstructed exactly with sinc interpolation. 
Computer graphics deals with multi-dimensional, multi-channel signals where the coordinates are often 2-dimensional, 3-dimensional, or higher and the output dimension is $d > 1$. The multi-dimensional axis can often be flattened into 1D and the channels can be dealt as separate signals.

A neural field~\cite{neuralfields2021} is a parametric function $\psi_\theta(x) \approx u(x)$ which approximates a continuous signal with a continuous function with parameters $\theta$, fitted through stochastic optimization.
The parameters of {\em Global} methods, which includes NeRF~\cite{mildenhall2020nerf}, consist entirely of the MLP's weights and biases.

\begin{wrapfigure}{r}{0.4\linewidth}
\vspace{-1mm}
\hspace*{-4mm}\begin{overpic}[width=1.05\linewidth]{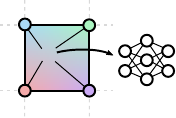}
\put(24.3,33.7) { $x$ }
\put(78,10) { $\psi_\theta$ }
\put(28,8) { $Z$ }
\put(15.3,58) { \small $\mathrm{interp}(x, Z)$ }
\end{overpic}
\vspace{-2mm}
\end{wrapfigure}

Conversely, {\em Feature-grid} methods augment the MLP with feature-grid parameters $Z$. The feature-grid is typically a regularly spaced grid, and the function $\mathrm{interp}$ is used to interpolate the local feature vectors $z = \mathrm{interp}(x, Z)$ for a given coordinate $x$. 

Since $\psi_\theta(x, \mathrm{interp}(x, Z)) \approx u(x)$ is a non-linear function, this approach has the potential to reconstruct signals with frequencies above the usual Nyquist limit. Thus coarser grids can be used, motivating their use in signal compression.

The feature grid can be represented as a matrix $Z \in \mathbb{R}^{m \times k}$ where $m$ is the number of grid points, and $k$ is the dimension of the feature vector at each grid point. Since $m \times k$ may be quite large compared to the size of the MLP, the feature vectors are by far the most memory hungry component. For an example, \citet{muller2022instant} utilize ten thousand MLP weights and \emph{12.6 million} feature grid parameters to represent radiance fields. We therefore wish to compress the feature grids and look to discrete signal compression for inspiration.

\begin{figure*}
    \small
    \begin{overpic}[width=\textwidth]{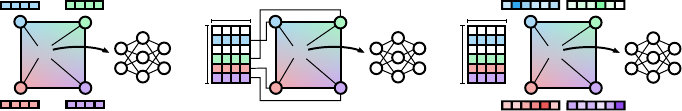}
        \put( 5.7, 8.2) { \large $x$ }
        \put(43.2, 8.2) { \large $x$ }
        \put(80.6, 8.2) { \large $x$ }
        \put( 2.0, -2.5) { \textbf{(a)} Feature grid }
        \put(16.0,  2.5) { MLP $\psi_\theta(x,z)$ }
        \put(33.0, -2.5) { \textbf{(b)} Index grid into feature codebook $D$ }
        \put(55.5,  2.5) { $\psi_\theta(x,z)$ }
        \put(27.9,  8.2) { $2^b$ }
        \put(32.8, 13.5) { $k$ }
        \put(32.8, 2.5) { $D$ }
        \put(67, -2.5) { \textbf{(c)} Soft-index grid. Differentiable lookup by ${C_i \cdot D}$ }
        \put(93.2,  2.5) { $\psi_\theta(x,z)$ }
        \put(65.4,  8.2) { $2^b$ }
        \put(70.5, 13.5) { $k$ }
        \put(70.5, 2.5) { $D$ }
        \put(74.6, 12.8) { $C_1$ }
        \put(74.6, 2.5) { $C_2$ }
        \put(88.4, 2.5) { $C_3$ }
        \put(88.4, 12.8) { $C_4$ }
    \end{overpic}
    \vspace{1mm}
    \caption{\label{fig:methodoverview}
        \textbf{(a)} shows the baseline uncompressed version of our data structure, in which we store the bulky feature vectors at every grid vertex, of which there may be millions. In \textbf{(b)}, we store a compact $b$-bit code per vertex, which indexes into a small codebook of feature vectors. 
        This reduces the total storage size, and this representation is directly used at inference time. 
        This indexing operation is not differentiable; at training time \textbf{(c)}, we replace the indices with vectors $C_i$ of width $2^b$, to which softmax $\sigma$ is applied before multiplying with the entire codebook. 
        This `soft-indexing' operation \emph{is} differentiable, and can be converted back to `hard' indices used in \textbf{(b)} through an {\em argmax} operation.
    }
\end{figure*}
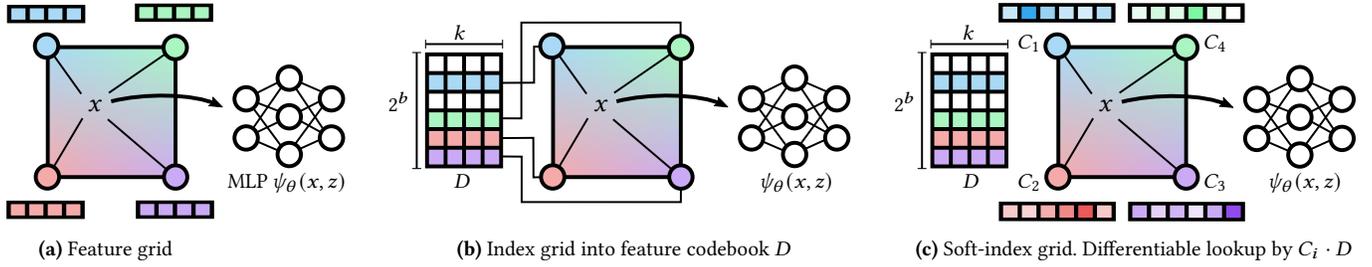

A standard method for compressing \textit{discrete} signals is known as transform coding~\cite{goyal2001theoretical},
where a function transforms the discrete signal $u_x$ to a representation $v_x$:
\begin{equation}
\begin{split}
    v_x &= f(u_x) \\
    u_x &= f^{-1}(v_x)
\end{split}
\end{equation}
We refer to the transformed representation $v_x$ as the \textit{transform coefficients}. The role of this transform is to decorrelate the signal such that quantization or truncation can be applied on the coefficients to effectively compress them. 

\textbf{\textit{Linear}} transform coding uses a linear transform $A \in \mathbb{R}^{n \times n}$ to produce the transform coefficients $v_x = A u_x$; the signal $u_x$ can then be reconstructed with the inverse $A^{-1}$. 
These transform matrices can be fixed (based on an efficient transform such as DFT and DCT~\cite{ahmed1974discrete}) or constructed from data. The Karhunen-Loève transform (KLT), for example, is a data-driven transform which can optimally decorrelate Gaussian distributed data.

\textbf{\textit{Non-linear}} transform coding~\cite{balle2020nonlinear} uses a parametric function $f_\gamma : \mathbb{R}^n \to \mathbb{R}^n$ with parameters $\gamma$ along with its inverse $f_{\gamma}^{-1}$ to encode and decode discrete signals. The transform $f_\gamma$ is often a neural network.
In comparison to the similar setup known as the \textit{auto-encoder}~\cite{kramer1991nonlinear, kingma2013auto}, non-linear transform coding has the additional goal of compressing the transform coefficients. 
On the other hand, \textit{auto-decoder}~\cite{park2019deepsdf} refers to only explicitly defining 
the inverse transform $f_\gamma^{-1}$ and performing the forward transform via stochastic optimization on the transform parameters $\gamma$ and the transform coefficients $v_x$. That is, the forward transform is $f(u_x) = \argmax_{v_x, \gamma} \| f_\gamma^{-1}(v_x) -  u_x \| $. 
This could also be seen as a form of stochastic variational inference~\cite{hoffman2013stochastic}. Similar to non-linear transform coding for auto-encoders, we define the \textit{compressed auto-decoder} as the compressive equivalent of the auto-decoder.

Computing the transform with respect to the entire sequence can be computationally expensive for large sequences. 
Block-based transform coding divides the signal $u_x$ into fixed size chunks of size $k$. Instead of computing global transform coefficients $v_x = A u_x$ with a large $n \times n$ matrix, we can reshape $u_x \in \mathbb{R}^n$ into a matrix $U \in \mathbb{R}^{\frac{n}{k} \times k}$. The smaller, \textit{block-wise} transform $\hat{A} \in \mathbb{R}^{k\times k}$ is applied to get block transform coefficients:
\begin{equation}
\begin{split}
    V & = U \hat{A} \\
    U & = V \hat{A}^{-1}
\end{split}
\end{equation}
In the non-linear case, we can use a function $f_\gamma : \mathbb{R}^k \to \mathbb{R}^k$ to code individual blocks (rows of the matrix $U$).
For further compression, the rows of $U$ can be clustered via vector quantization~\cite{gray1984vector}. 
We also compress our feature-grids using methods inspired by block-based compression, specifically vector quantization.

%% file: src/3_method.tex
\section{Method}
\label{sec:method}

We propose the \textit{vector-quantized auto-decoder} method which 
uses the auto-decoder framework with an extra focus on learning compressed representations. 
The key idea is to replace bulky feature-vectors with indices into a learned codebook\footnote{In prior work such as \cite{takikawa2021neural}, the feature vectors consumed 512 bits each; the codebook indices that replace them in this work may be as small as 4 bits.}.
These indices, the codebook, and a decoder MLP network are all trained jointly. 
See Fig. \ref{fig:methodoverview} for an overview of the method.

By eschewing the encoder function typically used in transform coding, we are able to learn compressed representations with respect to arbitrary domains, such as the continuous signal that a coordinate network MLP encodes, even under indirect supervision (such as training a neural radiance field from images with a volumetric renderer). We give an overview of the compressed auto-decoder framework in Section \ref{subsec:compressedautodecoder}, show how feature-grid compression fits into the framework in Section \ref{subsec:featuregridcomp}, and discuss our specific implementation of this framework 
in Section \ref{subsec:vqad}. 

\subsection{Compressed Auto-decoder}
\label{subsec:compressedautodecoder}

In order to effectively apply discrete signal compression to feature-grids,
we leverage the auto-decoder~\cite{park2019deepsdf} framework where only the decoder $f_\gamma^{-1}$ is explicitly constructed; performing the forward transform involves solving the following optimization problem through stochastic gradient descent:
\begin{equation}
\label{eq:autodecoder}
    \argmin_{v_x, \gamma} \| f_\gamma^{-1}(v_x) - u_x  \|.
\end{equation}

A strength of the auto-decoder is that it can reconstruct transform coefficients with respect to supervision in a domain different from the signal we wish to reconstruct. We define a differentiable forward map~\cite{neuralfields2021} as an operator $F$ which lifts a signal onto another domain. 
Now, we must solve the following problem:
\begin{equation}
\label{eq:forwardmap}
    \argmin_{v_x, \gamma} \| F( f_\gamma^{-1} (v_x)) - F(u_x)  \|
\end{equation}
For radiance field reconstruction, the signal of interest $u_x$ is volumetric density and plenoptic color, while the supervision is over 2D images. In this case, $F$ represents a differentiable renderer. 

\subsection{Feature-Grid Compression}
\label{subsec:featuregridcomp}

The feature grid is a matrix $Z \in \mathbb{R}^{m \times k}$ where $m$ is the size of the grid and $k$ is the feature vector dimension. Local embeddings are queried from the feature grid with interpolation at a coordinate $x$ and fed to a MLP $\psi$ to reconstruct continuous signals. The feature grid is learned by optimizing the following equation:
\begin{equation}
\label{eq:featuregrid}
    \argmin_{Z, \theta} \mathbb{E}_{x, y} \| F( \psi_\theta(x,  \mathrm{interp}(x, Z) ) ) - y \| 
\end{equation}
where $\mathrm{interp}$ represents trilinear interpolation of the 8 feature grid points surrounding $x$. 
The forward map $F$ is applied to the output of the MLP $\psi$; in our experiments, it is a differentiable renderer~\cite{mildenhall2020nerf} and $y$ are the training image pixels. 

The feature grid $Z$ can be treated as a block-based decomposition of the signal where each row vector (block) of size $k$ controls the local spatial region. Hence, we consider block-based inverse transforms $f_\gamma^{-1}$ with block coefficients $V$. Since we want to learn the compressed features $Z = f_\gamma^{-1}(V)$, we substitute $Z$:
\begin{equation}
\label{eq:compressedfeaturegrid}
    \argmin_{V, \theta, \gamma} \mathbb{E}_{x, y} \| F( \psi_\theta(x, \mathrm{interp}(x, f^{-1}_\gamma(V)))) - y \|.
\end{equation}

Considering the $F(\psi(x, \theta, \mathrm{interp}(x, Z)))$ as a map which lifts the discrete signal $Z$ to a continuous signal where the supervision (and other operations) are applied, we can see that this is equivalent to a block-based compressed auto-decoder. This allows us to work only with the discrete signal $Z$ to design a compressive inverse transform $f_\gamma^{-1}$ for the feature-grid $Z$, in our case the vector-quantized inverse transform to directly learn compressed representations.

\subsection{Vector-Quantization}
\label{subsec:vqad}
We show how vector quantization can be incorporated into the compressed auto-decoder framework.
We define our compressed representation $V$ as an integer vector $V~\in~\mathbb{Z}^{m}$ with the range $[0,~2^b-1]$. This is used as an \textit{index} into a codebook matrix $D~\in~\mathbb{R}^{2^b~\times~k}$ where $m$ is the number of grid points, $k$ is the feature vector dimension, and $b$ is the bitwidth. Concretely, we define our decoder function $f_D^{-1}(V) = D[V]$ where $[\cdot]$ is the indexing operation. See Fig. \ref{fig:methodoverview}\textbf{(b)}. The optimization problem is:
\begin{equation}
\label{eq:hardloss}
\begin{split}
    \argmin_{D, V, \theta} & \mathbb{E}_{x, y} \| \psi_\theta(x, \mathrm{interp}(x, D[V])) - y \| \\
\end{split}
\end{equation}
Solving this optimization problem is difficult because indexing is a non-differentiable operation with respect to the integer index $V$.

As a solution, in training we propose to represent the integer index with a \textit{softened} matrix $C \in \mathbb{R}^{m \times 2^b}$ from which the index vector $V = \argmax_i C[i]$ can be obtained from a row-wise argmax. We can then replace our index lookup with a simple matrix product and obtain the following optimization problem:  
\begin{equation}
\label{eq:softloss}
\begin{split}
    \argmin_{D, C, \theta} & \mathbb{E}_{x, y} \| \psi_\theta(x, \mathrm{interp}(x, \sigma(C)D)) - y \| \\
\end{split}
\end{equation}
where the softmax function $\sigma$ is applied row-wise on the matrix $C$. This optimization problem is now differentiable. (See Fig. \ref{fig:methodoverview}\textbf{(c)})

In practice, we adopt a straight-through estimator~\cite{bengio2013estimating} approach to make the loss be aware of the hard indexing during training. That is, we use Equation \ref{eq:hardloss} in the forward pass and Equation \ref{eq:softloss} in the backward pass. Other choices of approximations like the Gumbel softmax~\cite{jang2016categorical} exist, but we empirically find that straight-through softmax works well. 

At storage and inference, we discard the softened matrix $C$ and only store the integer vector $V$. Even without entropy coding, this gives us a compression ratio of $16 m k / (m b + k 2^b) $ which can be orders of magnitude when $b$ is small and $m$ is large. We generally observe $m$ to be in the order of millions, and evaluate $b \in \{4, 6\}$ for our experiments. In contrast to using a hash function~\cite{muller2022instant} for indexing, we need to store $b$-bit integers in the feature grid but we are able to use a much smaller codebook (table) due to the learned adaptivity of the indices.

Rather than a single resolution feature-grid, we arrange $V$ in a multi-resolution sparse octree as in NGLOD~\cite{takikawa2021neural}, to facilitate streaming level of detail. Thus, for a given coordinate, multiple feature vectors $z$ are obtained - one from each tree level - which can then be summed (i.e. in a Laplacian pyramid fashion) or concatenated before being passed to the MLP. We train a separate codebook for each level of the tree. Similarly to NGLOD~\cite{takikawa2021neural}, we also train multiple levels of details jointly.

%% file: src/4_results.tex
\section{Experiments}
\input{tables/baseline}
\input{tables/compression_vq}
\begin{figure}
  \includegraphics[width=\linewidth]{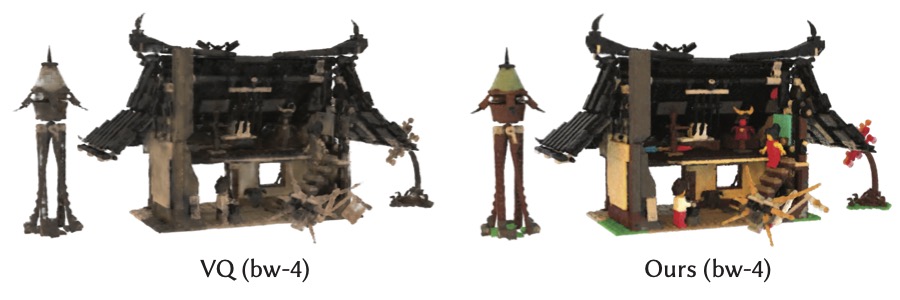}
  \caption{\textbf{Post-Process vs. Learned Vector Quantization}. We compare applying k-means vector quantization on the feature grid as a post-processing after training, vs. learning vector quantization end-to-end with the same number of codebook entries. We see the k-means quantization has visible discoloration, whereas ours preserves the visual quality.}
  \label{fig:learned_indices_vs_post}
\end{figure}

\begin{figure}
  \includegraphics[width=\linewidth]{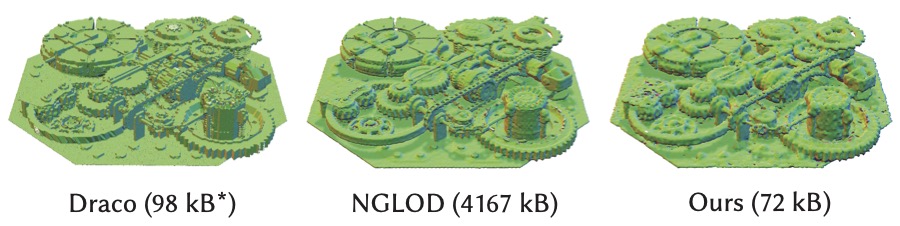}
  \caption{\textbf{Compressing geometry}. We show how VQ-AD can compress signed distance functions as in NGLOD. 
  Our method introduces visible artifacts in the normals, however it does result in a significant bitrate reduction. 
  We also compare against a quantized Draco mesh which has similar bitrates when entropy coded 
  (\SI{2}{\mega\byte} as the decompressed binary \tt{.ply} mesh).}
  \label{fig:sdf}
\end{figure}

\begin{figure}
  \includegraphics[width=\linewidth]{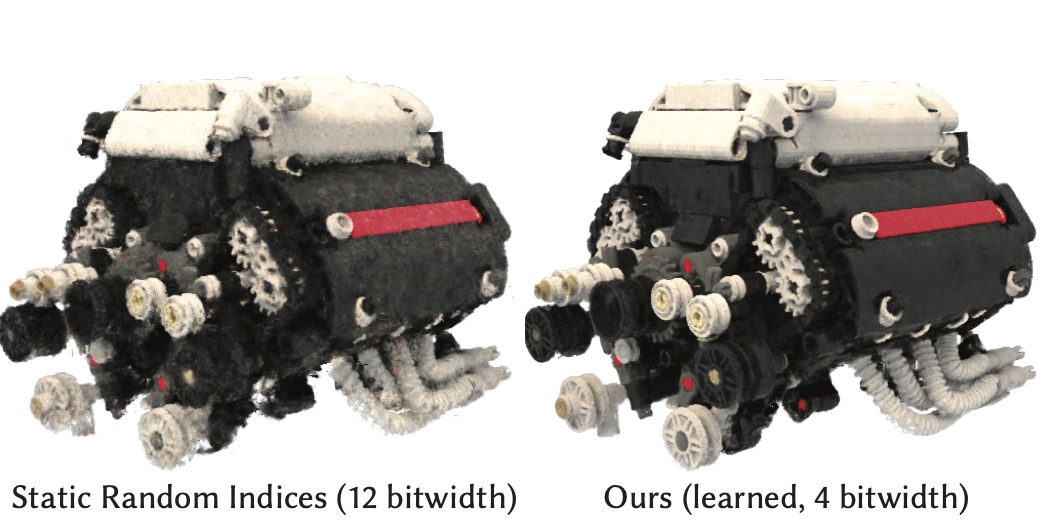}
  \caption{\textbf{Qualitative comparison of static and learned indices}. We qualitatively compare a hash approach with 12 bitwidth codebooks and our learned indices with 4 bitwidth codebooks which have similar compression rates. We see that our learned indices are able to reconstruct with less noise.}
  \label{fig:traditionalcompression}
\end{figure}

\input{tables/indices}

\begin{figure*}
  \includegraphics[width=\textwidth]{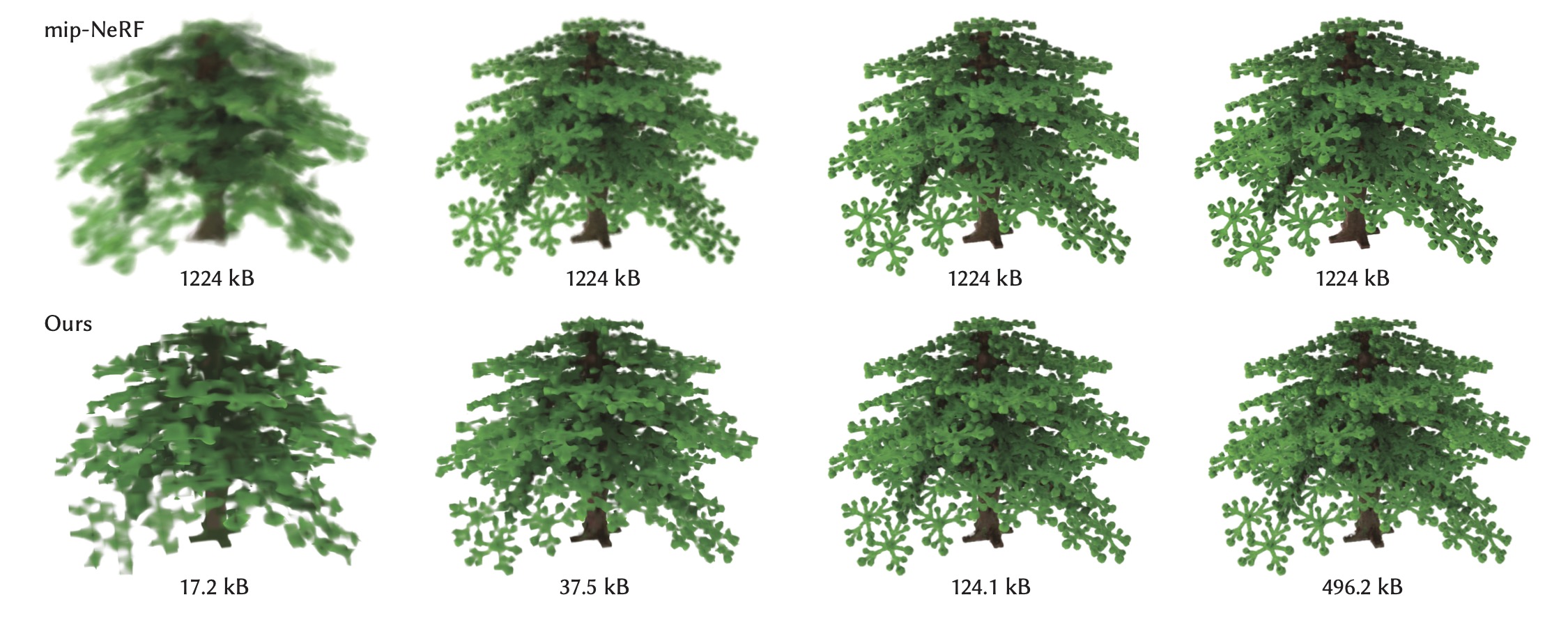}
  \caption{\textbf{Compressed levels of detail.} From left to right: the different mip levels. Top row: mip-NeRF at different cone widths. Although mip-NeRF produces filtered results, they are constant bitrate. Bottom row: Our multiresolution and vector quantized representation. We are able to simultaneously filter and compress the representation, making it suitable for progressive streaming and level of detail.
  }
  \label{fig:prefiltered}
\end{figure*}

\begin{figure}
  \includegraphics[width=\linewidth]{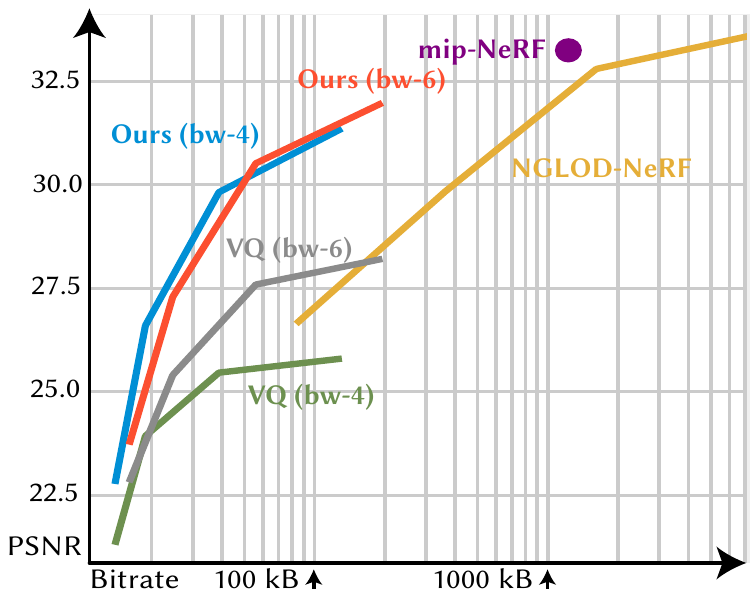}
  \caption{\textbf{Rate Distortion Curve}. This graph shows the rate-distortion tradeoffs of different methods on the `Night Fury' RTMV scene, 
  where the y-axis is PSNR and the x-axis is bitrate (in log-scale). 
  Single-bitrate architectures are represented with a dot. For Mip-NeRF (purple), the filtering mechanism can move the dot vertically, but not horizontally.
  Our compressed architecture (red and blue) has variable-bitrate and is able to dynamically scale the bitrate to different levels of details. 
  Our architecture is more compact than feature-grid methods like NGLOD (yellow) and achieves better quality than postprocessing methods like k-means VQ (gray and green).}
  \label{fig:ratedistortion}
\end{figure}

\subsection{Baseline and Implementation Details}

As an \textit{uncompressed} baseline for our experiments,
we implement the sparse multi-resolution feature grid architecture in NGLOD~\cite{takikawa2021neural} for the task of learning radiance fields from RGB-D input, with minor modifications to make NGLOD more suitable for this task (see supplemental materials for more details). Although we choose NGLOD as our baseline, our method is agnostic to the choice of data structure in which to store the feature grid.

We initialize NGLOD's octree with depth maps associated with each training image.
To train without such calibrated depth information, we could in principle 
adopt a coarse-to-fine approach~\cite{liu2020neural, yu2021plenoxels},
and refine the octree structure during training. However we do not evaluate this, choosing instead to focus on our proposed contributions in compression and streaming.
Note that the references we compare against did not use depth during training, and could benefit from depth supervision~\cite{kangle2021dsnerf}. 
These methods are mainly used only as reference points of quality. 

For all evaluations, we use the 10 brick scenes from RTMV dataset \cite{tremblay22rtmv} which has challenging high-complexity assets. 
We use the same evaluation scheme and evaluate the results at a $400 \times 400$ resolution with LPIPS~\cite{zhang2018unreasonable} (with VGG), SSIM~\cite{wang2004image}, and PSNR. 
We list detailed architectural hyperparameters in the supplemental material. 
For some figures, we use our own custom dataset of assets collected from TurboSquid which we renderered in NViSII~\cite{morrical2021nvisii}. 

Our reference comparisons with NeRF \cite{mildenhall2020nerf}, mip-NeRF~\cite{barron2021mip}, and Plenoxels~\cite{yu2021plenoxels} were produced using the author's code using default hyperparameters and with the same evaluation setting as described previously. 
Table \ref{tab:baseline} shows a comparison of the reconstruction quality of these prior methods to our own baseline, uncompressed NGLOD-NeRF. 
We are able to achieve comparable quality despite being orders of magnitudes faster than the global methods. 
The storage impact for the feature-grid methods is much higher than the global methods, motivating our compression technique which we evaluate next.

\subsection{Feature Grid Compression}

To evaluate the efficacy of the vector-quantized auto-decoder (VQ-AD) method, we evaluate several different baselines which perform compression as \textit{postprocessing} on a trained model. 
Our first baseline is low-rank approximation (LRA) through the KLT transform on individual feature vectors at different truncation sizes (f).
Our second baseline is vector quantization (kmVQ) through k-means clustering with different numbers of clusters at different quantization bitwidths (bw).
We report their average bitrate (see Table \ref{tab:classiccompression}) along with their compression rate with respect to the baseline. 
All results assume half-precision (16 bit) floating point numbers and do not use entropy coding. 
LRA is not competitive in quality nor compression ratio compared to the other methods. 
Figure~\ref{fig:learned_indices_vs_post} shows qualitative results between kmVQ and our method, where kmVQ causes noticeable discoloration.
This is corroborated by the quantitative results which show kmVQ to be at a significant quality disadvantage, at equal size; 
this shows the efficacy of learning the indices during training. 

Our VQ-AD method can also be applied in contexts other than fitting radiance fields. Figure~\ref{fig:sdf} shows the original results from NGLOD which fits truncated signed distance functions (TSDF) alongside the VQ-AD version. We also compare against Draco~\cite{googledraco}, which compresses meshes through entropy coding and heavy quantization of the vertex positions. Our method does introduce visible high frequency artifacts for TSDFs, but we are able to nonetheless reduce the bitrate significantly without relying on entropy coding.

\subsection{Random vs. Learned Indices}

Our codebook learning method can be seen as a special form of the hash encoding method from Müller et al.~\cite{muller2022instant} where instead of using a fixed hash function to index into the codebook, we learn the indices and bake them into the grid. Learning the indices allows adaptive collision resolution which in turn allows the use of much smaller codebook sizes at the cost of having to store indices.

To evaluate this tradeoff, we implement an equivalent method to the hash encodings in our NGLOD-NeRF baseline implementations by simply allocating random indices of range $[0, 2^b]$ with a corresponding codebook size. We train this for 120 epochs at several different bitwidths (bw), and show that learning the indices can use a much smaller bitwidth than in the random case (see Table~\ref{tab:vshash}) at approximately equal quality. 
Figure~\ref{fig:traditionalcompression} evaluates the visual quality of the random and learned approaches at roughly equal storage cost; we see that the random indices based approach has much more visible noise. 

\subsection{Streaming Level of Detail}

We also showcase our ability to learn compressed multi-resolution representations which can be used for progressive streaming. 
Figure \ref{fig:prefiltered} shows a visual comparison between the Fourier encoding-based filtering mechanism from mip-NeRF~\cite{barron2021mip} at different cone radii, in comparison to our multiresolution representation. 
Both are able to produce different levels of detail, but our method is able to also reduce the bitrate accordingly at lower resolutions thus enabling progressive streaming. 

Figure~\ref{fig:ratedistortion} shows the rate-distortion curves for different methods, including our compressed multi-resolution architecture. The graph shows that our VQ-AD can achieve orders of magnitudes smaller bitrates, without significantly sacrificing quality like post-processing methods, {\em e.g.,} kmVQ. 
The graph highlights that our representation has variable-bitrate and encodes multiple different resolutions which can be progressively streamed at different levels of detail.
The memory overhead of our method prevents us from evaluating higher bitrates and we hope to explore this frontier in future work.

%% file: tables/baseline.tex
\bgroup
\def\arraystretch{1.0}
\begin{table}[t]
\caption{\textbf{Baseline References}. This table shows the baseline feature-grid method (NGLOD-NeRF) in comparison to NeRF and mip-NeRF which are state-of-the-art global-methods, and Plenoxels which is also a feature-grid method. We see from the results that NGLOD-NeRF is a strong baseline with similar quality to both. All floats are half precision.} 
\begin{center}
\setlength{\tabcolsep}{6.2pt}
\rowcolors{2}{gray!10}{white}
\begin{tabularx}{\linewidth}{lcccr}
\toprule
\textit{Method} & \textit{PSNR} $\uparrow$ & \textit{SSIM} $\uparrow$ & \textit{LPIPS} $\downarrow$ & \textit{Storage [fp16]}   \\
\midrule
NeRF & 28.28 & 0.9398 & 0.0410 & 2.5 MB \\
mip-NeRF & 31.61 & 0.9582  & 0.0214 & 1.2 MB  \\
Plenoxels & 31.38 & 0.9617 & 0.0431 & $\approx$ 168 MB \\
\midrule
NGLOD-NeRF & 32.72 & 0.9700 & 0.0379 & $\approx$ 20 MB \\
\bottomrule
\end{tabularx}
\end{center}
\label{tab:baseline}
\end{table}
\egroup

%% file: tables/compression_vq.tex
\bgroup
\def\arraystretch{1.0}
\begin{table}[t]
\caption{\textbf{LRA, VQ vs loss-aware VQ (ours)}. 
This table shows the comparison between low-rank approximation (LRA), vector quantization (kmVQ) and learned vector quantization (ours) at different truncation sizes (for LRA) and different quantization bitwidths (for kmVQ and ours).
We see that across all metrics we see a significant improvement by learning vector quantization. The bitrate is data dependent, so we report average bitrate.} 
\begin{center}
\setlength{\tabcolsep}{9.8pt}
\rowcolors{2}{gray!10}{white}
\begin{tabularx}{\linewidth}{lccr}
\toprule
\textit{Method} & \textit{PSNR} $\uparrow$ & \textit{SSIM} $\uparrow$ & \textit{Bitrate (CR)}\\
\midrule
NGLOD-NeRF & 32.72 & 0.9700 & 20 MB (1.0$\times$) \\
\midrule
+ LRA (8f) & 29.09 & 0.9546 & 10.2 MB (2.0$\times$) \\
+ LRA (4f) & 26.98 & 0.9387 & 5.1 MB (4.0$\times$) \\
\midrule
+ kmVQ (6 bw) & 27.25 & 0.9322 & 0.49 MB (40.9$\times$) \\
+ kmVQ (4 bw) & 25.02 & 0.9112 & 0.33 MB (61.3$\times$) \\
\midrule
Ours (6 bw) & 30.76 & 0.9567  & 0.49 MB (40.9$\times$) \\
Ours (4 bw) & 30.09 & 0.9482  & 0.33 MB (61.3$\times$) \\
\bottomrule
\end{tabularx}
\end{center}
\label{tab:classiccompression}
\end{table}
\egroup

%% file: tables/indices.tex
\bgroup
\def\arraystretch{1.0}
\begin{table}[t]
\caption{\textbf{Comparison between random indices and learned indices}. This table shows the effects of learning codebook indices with VQAD at 120 epochs with different quantization bitwidths (bw). To highlight the tradeoff, we list the size of the indices $V$ and codebook $D$ separately. We see that even when storing indices, we are able to achieve higher quality than the hash-based approach.} 
\begin{center}
\setlength{\tabcolsep}{5.8pt}
\rowcolors{2}{gray!10}{white}
\begin{tabularx}{\linewidth}{lcccr}
\toprule
\textit{Method} & \textit{PSNR} $\uparrow$ &  $\|V\|$ & $\|D\|$ & \textit{Total BR (CR)}\\
\midrule
Hash (16 bw) & 29.75 & 0 kB & 8388 kB &  8400 kB (2.42$\times$) \\
Hash (14 bw) & 28.48 & 0 kB & 2097 kB & 2109 kB (9.65$\times$) \\
Hash (12 bw) & 26.66 & 0 kB & 524 kB & 536 kB (37.9$\times$) \\
Hash (10 bw) & 23.70 & 0 kB & 131 kB & 143 kB (141.9$\times$) \\
\midrule
Ours (6 bw) & 29.92 & 477 kB & 8 kB & 497 kB (40.9$\times$) \\
Ours (4 bw) & 29.60 & 318 kB & 2 kB & 332 kB (61.3$\times$) \\
Ours (2 bw) & 27.59 & 159 kB & 0.5 kB & 172 kB (118.5$\times$) \\
Ours (1 bw) & 25.57 & 79 kB & 0.3 kB & 92 kB (221.2$\times$) \\
\bottomrule
\end{tabularx}
\end{center}
\label{tab:vshash}
\end{table}
\egroup

%% file: src/5_conclusion.tex
\section{Conclusion}

Simultaneous filtering and compression is an important feature for real-life graphics systems. 
We believe that neural rendering~\cite{tewari2020state, tewari2021advances} and neural fields~\cite{neuralfields2021} will become more integrated into next generation 
graphics pipelines, and as such it is important to design neural representations that are able to perform the same signal processing operations currently possible with other representations like meshes and voxels.
We believe that our method, the vector-quantized auto-decoder, is a step forward in that direction as we demonstrated our method can learn a streamable, compressive representation with minimal visual quality loss.

One of the major drawbacks of our presented method is its training footprint in terms of memory and compute at training time, which requires the allocation of a matrix of size $m \times 2^b$ to hold the softmax coefficients before they are converted into indices at inference and storage. We believe that this could be addressed via a hybrid approach between random and learned indices, where instead of storing softened version of indices, we learn a parametric function with respect to coordinates which can predict softened indices on-the-fly.
Our approach is also directly compatible with highly efficient frameworks like instant neural graphics primitives~\cite{muller2022instant} and we believe that the synthesis of these techniques is a very exciting research direction.

%% file: src/6_supplemental.tex
\section{Implementation Details}

\subsection{Minor Modifications to NGLOD}

We follow the open source implementations NGLOD~\cite{takikawa2021neural} architecture available at \url{https://github.com/nv-tlabs/nglod}.
Since the original NGLOD architecture was designed for learning and rendering signed distance functions, we make minor modifications to make the architecture more suitable for learning neural radiance fields. 
First, in NGLOD, the features can only be queried in the regions where the sparse voxels are allocated for the given level. 
This can be an issue for radiance fields, because this leads to perturbations in the viewpoint causing rapid change in the voxels being traced, which can cause instability in training and rendering.
Instead, we modify the feature lookup function such that any location where sparse voxels are allocated for the \textit{coarsest} level in the multiresolution hierarchy can be sampled. If while traversing the tree for a location $x$ and the location is no longer allocated in the tree for finer resolutions, we simply return a vector of zeros for those levels.
In practice, the NGLOD architecture is a sparse Laplacian pyramid which sums the feature vectors from multiple levels, so the zeros end up being a no-op, allowing for an efficient implementation. We also use a single unified MLP which is shared across all levels instead of using a separate MLP per level as in the original NGLOD implementation. 

\subsection{Architectural Hyperparameters} 

We use a feature vector size of 16, concatenated with the 3-dimensional view direction which is positionally encoded to
produce an embedded view direction vector of size 27. The feature vectors are stored on a sparse, multi-resolution grid with resolutions $[2^5, 2^6, 2^7, 2^8]$.
The concatenated feature vector and embedded view direction create a 
vector of size 43, which is the input to the neural network. The neural network is a 2-layer network with a single hidden layer
of size 128 and an output dimension of 4 (density and RGB color). We use the ReLU activation function for the hidden layer, ReLU
activation on the density output, and sigmoid activation on the RGB color output. We initialize the feature grid with
normally distributed samples with a standard deviation of 0.01. Our implementation for volumetric integration 
uses 16 samples for each voxel that was intersected by each ray, and thus our implementation could benefit (in compute)
from an early termination scheme.
Because this creates a variable number of voxels and samples per ray, we cannot use standard PyTorch operations to integrate them, and as such we use custom packed CUDA primitives for volumetric integration to process them. 
We implemented everything with PyTorch~\cite{paszke2019pytorch}, custom CUDA kernels, and the differentiable rendering primitives from the Kaolin library~\cite{jatavallabhula2019kaolin}.

\subsection{Other training details}

The point cloud to initialize the sparse NGLOD grid is generated by taking the ray origins and ray directions for the
ground truth camera parameters and adding the directions multiplied by depth (pixel-wise) to the origins 
to produce a point cloud. We then normalize this point cloud within a normalize cube with range $[-1, 1]$. 
We apply this same normalization factors to the ray origins such that the cameras are aligned. We store
the normalization scale and offset in the model to use consistent offsets at validation time. 

All optimizations and evaluations are performed in SRGB space. We downscale the images using area-weighted 
bilinear interpolation implemented in OpenCV. On the ground truth images, we premultiply the alphas
to remove the boundary artifacts. Training on the baselines were trained for 600 epochs (unless otherwise noted)
with a batch size 
of 4096 with a learning rate of 0.001 with the Adam optimizer. 
We scale the learning rate of the feature grid by 100 which 
we find to be important for performance. We performed some minor experiments with TV regularizations as in Plenoxels~\cite{yu2021plenoxels}
however we found the effects to be minimal and as such we did not use them. 

To train multiple levels of details, we follow a similar strategy to NGLOD~\cite{takikawa2021neural} where we train multiple level of details with a single model. Whereas in NGLOD they sum the loss function from all levels and train them simultaneously, we instead randomly sample a level of detail per batch. We sample levels from a distribution where each level of detail (starting from the coarsest level) has a 2x more likely chance of being sampled compared to the previous level. We also find that \textit{only} sampling the finest level of detail also manages to learn some level of detail effects, although at compromised quality for the lower levels of detail.

\section{Other Experimental Details}

\subsection{Training and Inference Speeds}

Since the timings for both training and inference depends on the model, we will report timings for the Night Fury model on an RTX 8000. We make a note that we do not utilize any optimizations like early stopping which we expect will make a large impact on the training and inference performance.

\subsubsection{Inference} 

Inference runs at around 15 FPS at 720p with 8 GB memory for both the uncompressed NGLOD-NeRF baseline and our compressed version (4,6 bitwidth). We expect that the compressed version could be faster with an optimized implementation that utilizes cache better (as showcased by Instant NGP~\cite{muller2022instant}). These numbers are heavily influenced by viewpoints, batching, and other implementation choices. 

\subsubsection{Training}

Training for 600 epochs takes around 20 minutes for the uncompressed model and around 40 minutes for the compressed model. Both achieve reasonable PSNR (30+) within 50 epochs, which takes around 2 minutes for the uncompressed model and around 4 minutes for the compressed model. These numbers assume there is no extra logging, debug rendering, model checkpoint, etc happening.

The peak memory used during training is 8 GB for the uncompressed model, 8 GB for the 4 bitwidth compressed model, and 18 GB for the 6 bitwidth compressed model. While the memory usage in training is high (as noted in the limitation section of the paper), the memory usage for inference is not affected. 

\subsection{Entropy Coding}

In the experiments in the main paper, we do not use any entropy coding.
If we do use entropy coding (gzip) on the uncompressed NGLOD-NeRF weights, we get a 7\% reduction in size. Using entropy coding on the compressed weights yields a 4\% reduction in size. We generally find that the trained indices are somewhat uniformly distributed, leading to smaller gains made by entropy coding.

To make entropy coding more effective, we can apply entropy minimization on the softmax weights in training as a regularization. This can give up to a 56\% reduction in size through entropy coding, but at the cost of a large quality drop. Entropy coding also precludes streaming.